%% file: main.tex
\documentclass[letterpaper,twocolumn,10pt]{article}
\usepackage{usenix}

\usepackage{tikz}
\usepackage{amsmath}

\usepackage{filecontents}

\usepackage[nolist]{acronym} 

\usepackage{xcolor}

\usepackage{listings}
\usepackage{todonotes}
\usepackage{amsmath}
\usepackage{subcaption}
\usepackage{enumitem}

\usepackage[inkscapelatex=false]{svg}

\definecolor{codegreen}{rgb}{0,0.6,0}
\definecolor{codegray}{rgb}{0.5,0.5,0.5}
\definecolor{codepurple}{rgb}{0.58,0,0.82}
\definecolor{backcolour}{rgb}{0.95,0.95,0.92}

\definecolor{oliveGreen}{RGB}{22, 139, 22}

\newcommand{\expertune}[0]{\textsc{MoETuner}}

\newcommand{\niparagraph}[1]{\vspace{0pt}\noindent\textbf{#1}}

\begin{document}

\date{}

\title{\Large \bf\textsc{MoETuner}: Optimized Mixture of Expert Serving with Balanced \\ Expert Placement and Token Routing}

\author{
{\rm Seokjin Go}\\
seokjin.go@gatech.edu\\
Georgia Institute of Technology
\and \and
{\rm Divya Mahajan}\\
divya.mahajan@gatech.edu\\
Georgia Institute of Technology
} 

\maketitle

\input{acronym.tex}
\input{./body/abstract}
\input{./body/introduction}
\input{./body/background}
\input{./body/motivation}
\input{./body/methodology}
\input{./body/evaluation}

\input{./body/conclusion}
\bibliographystyle{plain}
\bibliography{main}

\end{document}

%% file: acronym.tex
\begin{acronym}
\acro{Title}{ExPerTune}
\end{acronym}

%% file: body/abstract.tex
\begin{abstract}
Mixture-of-Experts (MoE) model architecture has emerged as a promising solution for scaling transformer models efficiently, offering sparse activation that reduces computational costs while increasing model capacity. 
MoE models activate only specific experts per token, unlike dense models that utilize the entire model for every token. This design allows for an increase in the total number of parameters through the inclusion of multiple experts, without a corresponding increase in computational intensity.
However, as MoE models scale, they need to be distributed across GPU devices, thus face critical performance bottlenecks due to their large memory footprint.
Expert parallelism distributes experts across GPUs, however, faces key challenges including an unbalanced token routing and expert activation, resulting in communication tail latency and processing inefficiencies. 
While existing solutions address some of these issues, they fail to resolve the dual challenges of load imbalance and communication skew.
The imbalance in token processing load across experts causes uneven processing times on different GPUs, while communication skew between GPUs leads to unbalanced inter-GPU data transfers. These factors degrade the performance of MoE models by increasing tail latency and reducing overall throughput.
To address these limitations, we propose an Integer Linear Programming (ILP) formulation to optimize expert placement by jointly considering token load, communication, and computation costs. 
We exploit the property that there is a token routing dependency across layers, where tokens routed to a specific expert in one layer are likely to be routed to a limited set of experts in the subsequent layer.
Our solution, \expertune, offers an optimal expert-to-GPU assignment that minimizes inter-GPU token routing costs and balances token processing across devices, thereby reducing tail latency and end-to-end execution time.
By resolving these systemic inefficiencies, our method significantly enhances the performance and efficiency of MoE models in a distributed setting. 
Experimental results demonstrate 9.3\% and 17.5\% of end-to-end speedups for single-node and multi-node inference respectively, showcasing the potential of our ILP-based optimization for offering expert parallel solutions for next-generation MoEs. 

\end{abstract}

%% file: body/introduction.tex
\section{Introduction}
The rapid growth of transformer models~\cite{bert,gpt2,gpt3,gemma,llama,t5} has revolutionized deep learning, driving advancements in natural language processing, computer vision, and other sequence-based tasks.
By leveraging self-attention mechanisms, transformers excel at capturing long-range dependencies and processing large-scale datasets efficiently.
However, as these models scale, their parameter count grows exponentially, significantly increasing computational and memory demands.
For instance, models like LLaMA and GPT, with billions of parameters, push the limits of current hardware, making large-scale training and deployment prohibitively expensive~\cite{llama, gpt3}.

As such, Mixture-of-Experts (MoE) architecture has emerged to efficiently scale large models~\cite{mixtral, switchtransformer, gshard, cai2024survey}. 
MoE models activate only a subset of experts within each layer, significantly reducing the computational burden compared to dense models~\cite{shazeer2017outrageously}.
By routing tokens through only a subset of experts at each layer, MoE models reduce computational costs compared to dense models of equivalent capacity.
This sparse activation allow for larger models to be trained without a proportional increase in compute requirements. 
%
Despite these benefits, MoE models face several challenges as they continue to scale. 
While sparse routing helps reduce compute intensity, the parameters still scale and so does the memory capacity requirement~\cite{mcsmoe,deepspeed, switchtransformer, fastermoe, fastmoe, cai2024shortcut,hwang2024pre}.
Additionally, the effectiveness of individual experts becomes increasingly unbalanced, with some experts being activated far more frequently than others. 
%
%
This skewed activation pattern results in imbalanced hardware utilization of MoE models.

\begin{figure}
    \centering
    \includesvg[width=0.85\linewidth]{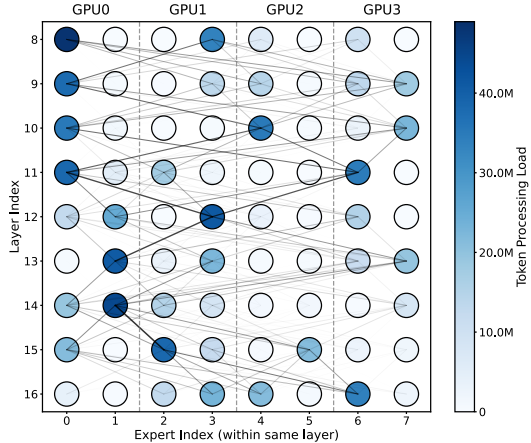}
    \vspace{-1em}
    \caption{Token routing statistics for Mixtral-8x7B. Each colored circle represents an expert in a layer, and the lines connecting them illustrate the number of tokens routed between pairs of experts. Thicker lines indicate a higher volume of routed tokens, highlighting key routing dependencies.
    }
    \label{fig:intro-tokenrouting}
\end{figure}

Expert parallelism distributes experts across multiple GPUs to scale the experts and size of the models~\cite{deepspeed-moe, exflow,  li2023accelerating, fastermoe, fastmoe,cai2024shortcut, singh2023hybrid}.
In a typical MoE model, each expert is responsible for processing a subset of tokens. 
As the number of experts grows, it becomes impractical to fit all of them onto a single GPU due to memory constraints. 
Expert parallelism solves this issue by partitioning the experts evenly across GPUs, ensuring that each GPU processes a smaller subset of experts~\cite{deepspeed-moe, li2023accelerating}.
In conventional expert parallelism implemented in popular ML frameworks like Deepspeed and Megatron-LM~\cite{deepspeed,megatron}, each GPU is assigned a contiguous range of experts.
For instance, when expert parallelism is deployed across 4 GPUs for an MoE model with 8 experts per layer, as shown in Figure~\ref{fig:intro-tokenrouting}, GPU 0 handles experts 0 and 1, GPU 1 handles experts 2 and 3, and so on.
In this setup, tokens are dispatched to remote GPUs via all-to-all communication if the destination expert is not present locally. 
This distribution aims to alleviate the memory burden on individual GPUs, enabling the model to scale efficiently as the number of experts increases.

Expert parallelism introduces bottlenecks. 
Firstly, communication during token routing, especially all-to-all, can dominate execution time, especially when token routing is imbalanced across GPU pairs. 
For example, for the Mixtral-8x7B~\cite{mixtral} model, we observe that all-to-all communication takes up 35.7\% of the end-to-end inference time.
Secondly, in existing work~\cite{exflow,li2023accelerating, fastermoe, singh2023hybrid} even though every GPU has equal number of experts, the expert activation is still skewed and activates certain experts more than the others. 
Uneven activation of experts across GPUs exacerbates token processing tail latency, where GPUs responsible for processing a larger number of tokens cause other GPUs to stall until all computations are finished. 
In summary, while expert parallelism mitigates memory constraints to facilitate scalability, it adds the communication overhead and faces token balance issues.

To address inefficiencies in expert parallelism, prior works explore techniques such as overlapping compute and communication to mitigate communication costs~\cite{deepspeed-moe, megatron, li2023accelerating, fastermoe,cai2024shortcut,singh2023hybrid,jiang2024lancet}.
Other approaches focus on minimizing communication volume through optimized token dispatch mechanisms or locality-aware expert placement strategies, achieving notable gains in throughput and efficiency~\cite{exflow}. 
While these methods achieve notable improvements in throughput and efficiency by reducing the overhead of inter-GPU communication, they often fall short in addressing two critical challenges: load imbalance in token processing and the communication tail latency caused by uneven token distribution across GPU pairs. 
These works still distribute the experts evenly across GPUs purely based on parameter size and not based on their token activation. 
These systemic inefficiencies remain significant bottlenecks, preventing optimal scalability and performance in large-scale MoE systems.

To address these challenges, we leverage the insight that activation dependencies exist between experts across layers, creating an affinity for activating specific experts based on those activated in the previous layer.
Moreover, certain experts are activated significantly more frequently than others, adding an imbalance to the system.
A straightforward approach is to distribute experts based on token routing; however, the challenge with this approach is that distributing with expert parallelism still needs to mitigate the memory bottleneck, thus parameter size is still a factor that determines how experts are distributed.
In this work, instead of distributing experts evenly based solely on parameter size, we account for size, token routing patterns, and expert activation chains.
To solve this optimization problem, we propose an Integer Linear Programming (ILP) formulation that models the communication and computation costs of token routing and expert activation.
By leveraging cross-layer dependencies, our approach derives an optimal expert placement strategy that minimizes inter-GPU communication while balancing token processing workloads and expert size across devices.
This strategy effectively reduces all-to-all communication overhead and mitigates compute-induced tail latency by ensuring even distribution of token dispatching and expert activations, while minimizing the total number of tokens dispatched.
Through ILP-based optimization, we resolve communication skew and token processing load imbalances, ultimately enhancing the scalability and performance of MoE models.

In summary, we make the following contributions:
\begin{itemize}
    \item We identify key bottlenecks in existing expert parallelism techniques, which focus primarily on reducing total communication volume but often result in imbalanced token routing and uneven expert activation across devices.
    \vspace{-0.5em}
    \item We exploit token routing dependencies to design an optimal expert placement strategy that mitigates these bottlenecks, improving load balancing and communication efficiency across GPUs.
    \vspace{-0.5em}
    \item We introduce \expertune~, a novel optimization approach for expert parallelism in MoE models leveraging ILP to minimize both communication and compute-induced tail latency.
    \vspace{-0.5em}
\end{itemize}

We demonstrate the effectiveness of \expertune~using the Mixtral-8x7B model, achieving 9.3\% and 17.5\% of average speedups in single-node (8 H100 GPUs) and multi-node (16 H200 GPUs) distributed MoE inference tasks, respectively.
These improvements are driven by two key factors: (1) load-balanced token processing to minimize GPU idle times and lower token processing latency, and (2) reduced variation in remote token dispatching, ensuring efficient all-to-all communication.
Together, these optimizations address inefficiencies in expert computation and token dispatching, enabling consistent performance improvements across distributed settings.

%% file: body/background.tex
\section{Background}\label{sec:background}

\subsection{Mixture-of-Experts (MoE)}
Mixture-of-Experts (MoE) models have emerged as a powerful approach for scaling deep learning architectures by partitioning model parameters into specialized subsets, known as “experts”, which process tokens selectively~\cite{gemma, switchtransformer, mixtral,gshard,shazeer2017outrageously,deepspeed-moe}. 
This selective routing of tokens to specific experts enables MoE models to achieve high model capacity without a proportional increase in computational costs, making them particularly effective for diverse tasks and datasets.
In MoE models, a sparse routing mechanism directs each token to a small fraction of experts—typically one or two per token~\cite{switchtransformer,gshard}.
This selective activation allows the model to maintain a low computational footprint while preserving the expressiveness of larger architectures, as only a subset of parameters is activated for each input. 
Consequently, MoE models are able to optimize both memory and compute efficiency per token processing, as only the necessary parameters are involved in the processing of each token.

The structure of an MoE model typically involves feed-forward layers interspersed with a router that determines which experts to activate based on token characteristics. 
Some models, such as Mixtral, employ a top-2 router that routes each token to the two most relevant experts, balancing model versatility and computational cost~\cite{mixtral}.
This routing design allows the model to adaptively distribute workload among experts, enhancing both accuracy and efficiency across varied inputs. 
%
%
Despite these advantages, MoE models face significant challenges in deployment as the number of experts increases.
While selective activation mitigates computational costs, the excessively scaled number of parameters can result in memory capacity constraints~\cite{mcsmoe,hwang2024pre,huang2023towards,kim2021scalable}.
These memory constraints necessitate the use of parallelization strategies to distribute both model parameters and computation across devices.
%
%

%
%

\subsection{Expert Parallelism and Token Dispatching}
\begin{figure}[h!]
    \centering
    \begin{subfigure}[t]{0.36\linewidth}
        \centering
        \includegraphics[width=\linewidth]{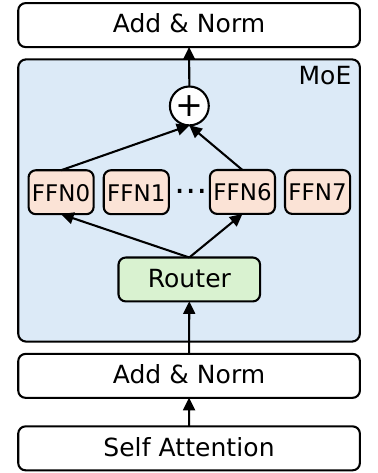}
        \caption{Execution of an MoE model on a single GPU with 8 experts. }
        \label{fig:moe-single-gpu}
    \end{subfigure}
    \hfill
    \begin{subfigure}[t]{0.54\linewidth}
        \centering
        \includegraphics[width=\linewidth]{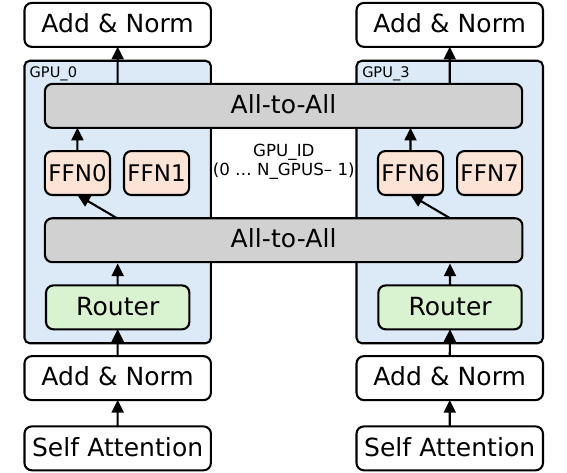}
        \caption{Expert-parallel execution with an expert parallel size of 4. Experts are distributed across GPUs, with all-to-all operations before and after the FFN computations.}
        \label{fig:moe-expert-parallel}
    \end{subfigure}
    \vspace{-0.8em}
    \caption{An example of Mixture-of-Experts (MoE) model execution. (a) Single-GPU execution with all experts local to the GPU. (b) Expert-parallel execution, where experts are distributed across GPUs, requiring inter-GPU communication through all-to-all operations.}
    \label{fig:background-ep}
\end{figure}

To manage the massive parameter size of MoE models, distributed systems employ expert parallelism—a model-parallel strategy designed specifically for MoE architectures.
Expert parallelism partitions experts across multiple GPUs, distributing computational workloads and reducing the memory footprint on each GPU.
This approach enables the model to scale without requiring an impractically large amount of memory on individual GPUs.
There are two primary techniques for distributing experts across devices to reduce their memory footprint. The first evenly splits the number of experts across devices~\cite{deepspeed, deepspeed-moe, megatron, exflow}, while the second splits each expert and distributes its components across devices~\cite{vllm, megatron,singh2023hybrid}.
Figure~\ref{fig:background-ep}(a) illustrates an example MoE model on a single GPU, while (b) demonstrates expert parallel execution across four devices, which involves all-to-all communication~\cite{nccl} to gather the expert data.
In the second expert parallel setup, tokens are routed to remote GPUs using all-to-all communication based on their assigned experts. This mechanism ensures that tokens are processed by the most relevant experts.
Despite these strategies, both techniques distribute experts evenly based solely on memory footprint, without accounting for the computational load on each expert.

\vspace{-1em}


\subsection{Related Work}

\niparagraph{Distributing deep learning models across devices.}
Distributing deep learning models across multiple devices has been a common strategy to overcome memory and computational limitations. 
Data parallelism~\cite{li2020pytorch,zhao2023pytorch} replicates the model across devices, with each device processing a subset of the input data independently and synchronizing gradients during updates.
Pipeline parallelism~\cite{huang2019gpipe,narayanan2019pipedream,narayanan2021memory} partitions a model into sequential stages, assigning each stage to a different device, and enables concurrent processing of inputs for different batches.
Model parallelism~\cite{megatron, switchtransformer}, unlike data and pipeline parallelism, partitions the model itself across multiple devices.
For instance, in tensor parallelism, large tensor computations are split into smaller sub-operations distributed across devices. 
Each device computes a portion of the operation, exchanging intermediate results with others during the forward and backward passes. 
This approach enables running models that exceed the memory capacity of a single device but introduces inter-device communication overhead, particularly for large matrix multiplications.
Expert parallelism~\cite{switchtransformer,deepspeed-moe,cai2024shortcut,chen2022ta} is a form of model parallelism tailored for MoE models, where the experts are distributed across devices.

\niparagraph{System level optimizations for MoE.}
To enhance scalability and efficiency in distributed training and inference of MoE models, various system level optimizations have been introduced.
DeepSpeed-MoE~\cite{deepspeed-moe} integrates multidimensional parallelism and hierarchical all-to-all algorithm to improve the scalability of distributed MoE inference.
Tutel~\cite{hwang2023tutel} incorporates adaptive parallelism and pipelining optimization during runtime to accommodate the dynamic nature of MoE models.
Other works, such as Switch Transformers~\cite{switchtransformer}, reduce the number of active experts per input to simplify routing and lower memory overhead. 
However, this often exacerbates token routing imbalances and leads to suboptimal resource utilization in distributed settings.
While these approaches aim to improve throughput and reduce resource overhead, they often fail to address critical challenges such as activation sparsity overheads and the load imbalances inherent in MoE models.
As a result, critical bottlenecks like inter-GPU communication and workload imbalances caused by real-world token routing patterns remain unaddressed, limiting their effectiveness in large-scale deployments.

\niparagraph{Expert parallelism to mitigate the memory bottleneck of MoE models.}
Several prior works have addressed the challenges of optimizing expert parallelism through techniques that focus on overlapping expert computation with communication or reducing the total number of communication steps.
%
Lancet\cite{jiang2024lancet} improves MoE performance by overlapping non-MoE computations with all-to-all through careful partitioning and pipelining of the training graph.
FasterMoE\cite{fastermoe} introduces a congestion-avoiding expert selection strategy that dynamically adjusts token assignments to relieve network congestion, thereby reducing training latency in distributed MoE models. 
Lina\cite{li2023accelerating} employs all-to-all prioritization and dynamic resource scheduling to mitigate communication bottlenecks and reduce resource contention during both training and inference. 
ExFlow~\cite{exflow} exploits token routing dependencies between different layers by strategically placing experts with the highest interdependencies on the same GPU. 
These works often aim to maximize GPU utilization by overlapping the computation of one expert with the communication for another, or by reducing the communication frequency through techniques like pipelining. 
However, they tend to overlook the impact of communication imbalance on tail latency, which becomes increasingly important in large-scale MoE models. 
%

In contrast, our approach specifically targets tail latency by formulating an ILP-based optimization strategy that balances both computation and communication loads. By considering token processing and communication overhead together, our method ensures more efficient scaling and improved performance under distributed conditions.

%% file: body/motivation.tex
\section{Challenges and Motivation}

\begin{figure}
    \centering
    \includesvg[width=1.0\linewidth]{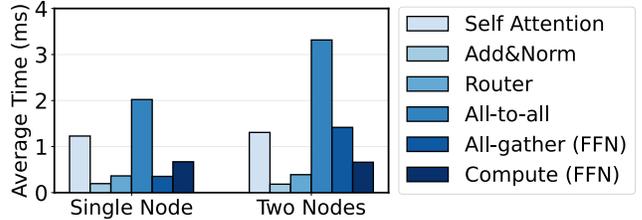}
    \vspace{-2em}
    \caption{Time distribution of representative operations during the forward pass of Mixtral-8x7B. The inference time is primarily dominated by all-to-all communication between GPU pairs, particularly in multi-node environments.}
    \label{fig:time-distribution}
    \vspace{-2ex}
\end{figure}

As we scale MoE models to include more experts, several challenges arise, affecting both memory and communication efficiency.
While increasing the number of experts enhances model capacity, it also significantly raises the memory footprint.
To address this, MoE models are distributed across multiple devices to alleviate memory constraints.
State-of-the-art expert parallelism techniques have been developed to manage these memory bottlenecks~\cite{fastermoe,deepspeed-moe,li2023accelerating,exflow}.
%
%
%
While these works mitigate compute and communication bottlenecks through optimized token dispatch mechanisms or locality-aware expert placement, they still suffer from inherent load imbalance and communication tail latency.
Expanding the number of experts exacerbates inter-GPU communication, as only a small subset of tokens is typically processed by local experts on a given GPU, requiring frequent token dispatches to remote GPUs.
In distributed multi-node environments, communication costs often surpass computation time, with all-to-all token routing causing network congestion and bandwidth limitations.
Figure~\ref{fig:time-distribution} illustrates the time distribution of representative operations during the forward pass of an MoE model, highlighting that communication time dominates as the system scales to multiple nodes.
Maintaining balanced loads across GPUs is crucial in large-scale deployments, as skewed token distributions can lead to significant delays, reduced throughput, and inefficient resource utilization.

\subsection{Challenges with Expert Parallelism}

\begin{figure}
    \centering
    \includesvg[width=1.0\linewidth]{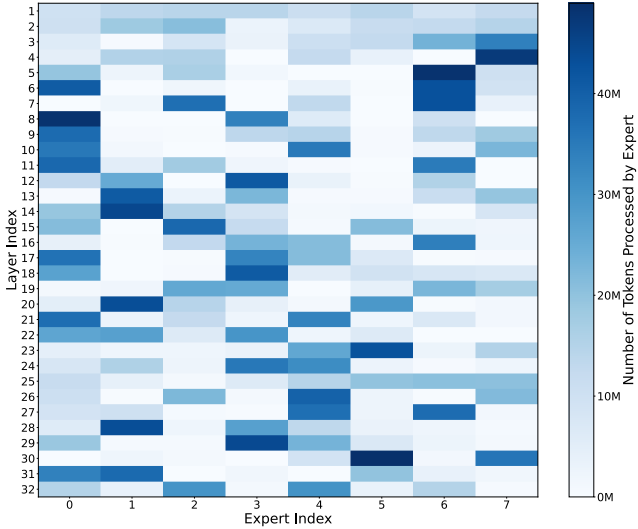}
    \vspace{-1.5em}
    \caption{Expert activation frequency of Mixtral-8x7B, highlighting significant load imbalance across layers. Darker regions indicate a higher skew. For example, in layer 14, experts 0 and 1 process 64\% of the total tokens. 
    }
    \label{fig:expert-load-imbalance}
    \vspace{-2ex}
\end{figure}

\niparagraph{Expert load imbalance.}
A significant challenge in distributing experts across devices is the occurrence of load imbalance, driven by skewed token routing distributions.
Current approaches distribute experts primarily based on memory requirements, often neglecting the compute load, which depends on the number of tokens processed by each expert.
This can result in certain experts receiving a disproportionate share of tokens, causing the GPUs to host these experts and experience higher processing loads. The imbalance leads to higher tail latencies, where some GPUs complete their computations early and remain idle while waiting for others to finish.
Figure~\ref{fig:expert-load-imbalance} highlights the expert activation frequency across layers, demonstrating a clear skew in the number of tokens processed per expert.
We observe that in layers with high skew, certain GPUs suffer from a significantly higher token processing load than others.
For instance, in layers 14 and 23, GPUs 0 and 3 process over 64\% and 69\% of the total tokens in each layer, respectively.
These inefficiencies are especially detrimental in high-throughput scenarios, as idle GPU time directly reduces overall throughput and resource utilization.
Addressing this load imbalance is essential to ensure efficient MoE processing and maximize hardware utilization.

\noindent\fbox{%
    \parbox{\columnwidth}{%
       \textbf{Challenge: Unbalanced token processing load across GPUs.} Token distribution across GPUs can become uneven due to imbalances in the number of tokens routed to different experts. Prior work only distributes experts to alleviate the memory footprint, this results in some GPUs becoming idle while others are overburdened due to token processing skew, leading to increased processing latency and reduced overall throughput.}
       }

\noindent\fbox{%
    \parbox{\columnwidth}{%
       \textbf{Insight: Token Routing-Based Expert Placement.} Strategic expert placement, rather than solely focusing on memory size mitigation, can significantly reduce load imbalance by evenly redistributing token workloads across GPUs in each layer.
        By maximizing GPU utilization and minimizing idle times, this approach effectively reduces token processing latencies, improving MoE model performance in both single-node and multi-node configurations.}%
}
\vspace{0.5em}

\begin{figure}[h!]
    \centering
    \includesvg[width=1.0\linewidth]{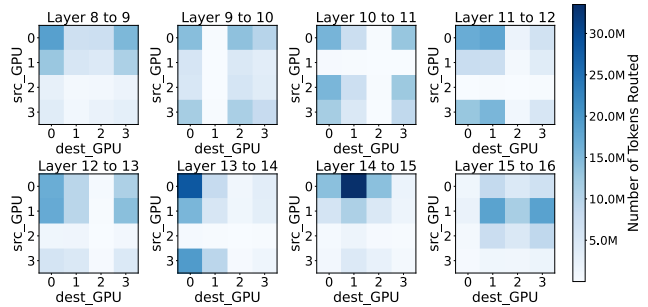}
    \vspace{-1.5em}
    \caption{Number of tokens dispatched across different GPU pairs. Certain GPU pairs experience substantially higher communication volumes compared to others.}
    \label{fig:token-dispatching}
\end{figure}

\niparagraph{Tail latency due to skewed token load and inter-GPU communication.}
Due to the distribution of experts across GPUs, inter-GPU communication introduces significant overhead.
Efficient inter-GPU communication is a critical challenge in expert-parallel MoE systems, as poorly planned expert placement can lead to communication tail latencies.
Frequent routing of tokens between remote GPUs, caused by suboptimal expert distribution, generates heavy all-to-all traffic patterns.
This issue is further exacerbated by token processing imbalances, which amplify communication skews across GPU pairs.
As a result, certain GPU pairs handle significantly higher communication volumes, leading to underutilized interconnect bandwidth and increased tail latencies, as other GPUs stall while waiting for communication operations to complete.
Figure~\ref{fig:token-dispatching} illustrates the disparity in token dispatching across GPU pairs, highlighting how some GPUs bear a disproportionately large communication load due to imbalanced token routing among distributed experts.

\noindent\fbox{%
    \parbox{\columnwidth}{%
\noindent \textbf{Challenge: Skewed Communication Patterns.} Routing tokens to experts on different GPUs generates substantial all-to-all communication. Without optimization, this can cause congestion and inefficient interconnect bandwidth utilization, ultimately limiting system scalability and performance.}}

\noindent\fbox{%
    \parbox{\columnwidth}{%
\noindent \textbf{Insight: Affinity Towards Certain Experts Across Layers.} Optimizing expert placement to account for inter-expert token routing patterns, remote token routing can be minimized, and communication loads can be distributed more evenly across GPU pairs.
This approach improves interconnect resource utilization, reduces latency, and ensures more predictable and efficient communication in large-scale MoE deployments.}}

\subsection{Inter-layer Token Routing Dependency}

While expert parallelism offers a scalable approach for deploying MoE models across multiple GPUs, its effectiveness hinges on efficient management of both processing load and inter-GPU communication.
Achieving this balance necessitates optimization of expert placement and token routing mechanisms.
To address these challenges, we exploit the observation that inter-layer expert token routing exhibits significant dependencies: when a token is routed to a specific expert in layer  \( l \), it is more likely to be routed to certain experts in layer \( l+1 \), and these patterns often persist across subsequent layers.
This reveals that token paths across layers are not random but instead follow structured and predictable patterns, driven by the underlying model dynamics and characteristics.
%

\begin{figure}[h!]
    \centering
    \includesvg[width=1.0\linewidth]{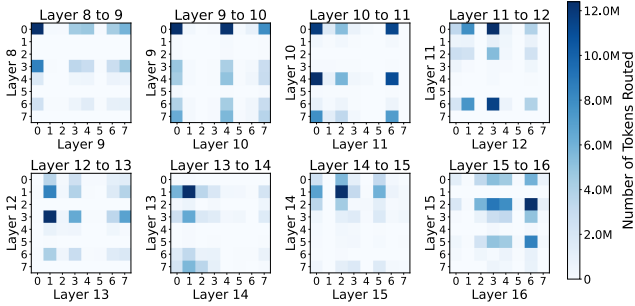}
    \vspace{-1.5em}
    \caption{Dependency table of Mixtral-8x7B illustrating inter-layer routing dependencies during inference. Most tokens routed to specific experts in one layer are directed to particular experts in the next layer. For instance, tokens processed by expert 1 at layer 14 are predominantly routed to expert 2 at layer 15, highlighting strong routing dependencies.}
    \label{fig:expert-dependency-table}
    \vspace{-3ex}
\end{figure}

Figure~\ref{fig:intro-tokenrouting} and~\ref{fig:expert-dependency-table} illustrate inter-layer routing dependencies, showing the frequency of token transitions between specific experts across layers.
These structured dependency chains provide a key opportunity for optimization. By aligning the physical placement of frequently interacting experts with these routing patterns, we can significantly reduce inter-GPU communication, improve load balance, and minimize latency across large-scale deployments.
This insight motivates the need for a global expert placement framework that goes beyond just alleviating memory overheads or being limited to layer-by-layer optimization.
By considering the holistic routing patterns across the model, such a framework ensures more efficient utilization of both computational and communication resources.
This approach directly tackles both the challenges of inter-GPU communication overhead and load imbalance, enabling more efficient and scalable MoE systems.

%% file: body/methodology.tex
\section{\expertune~for Expert Placement}

\begin{figure*}
    \centering
    \includegraphics[width=0.8\linewidth]{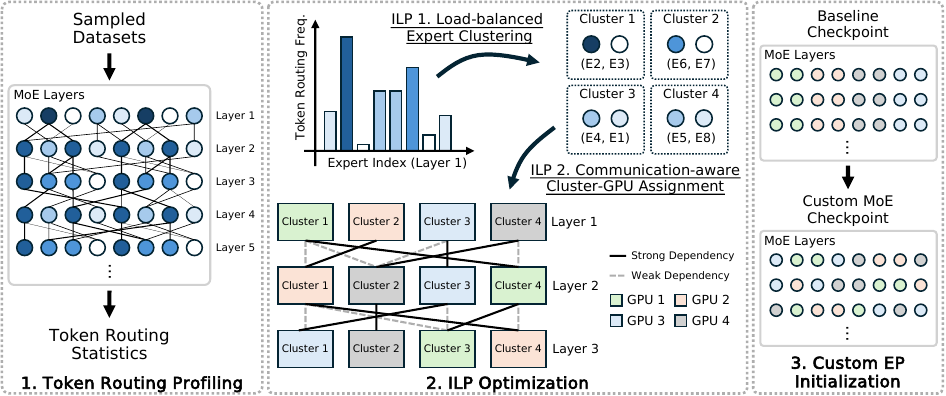}
    \caption{Overview of the \expertune~framework.}
    \label{fig:methodology-overview}
    \vspace{-3ex}
\end{figure*}

Based on the observations presented in Section \ref{sec:background}, we propose an \expertune~ framework a novel way to optimize expert placement for MoE models. This framework is built on the insight that inter-layer token routing exhibits predictable dependencies.
The primary goal of \expertune~is to develop an expert placement strategy that minimizes two critical factors: the imbalance of token processing load across GPUs and the inter-GPU communication overhead.
To leverage the expert routing dependency, we need to determine the routing per dataset.
However, using the entire dataset to determine the token routing can be prohibitive. 
Instead, we observe that we can perform inference on a small subset of the dataset for each task over a predefined number of iterations.
Our experiments in Section~\ref{subsec:designoverview} reveal that the routing patterns of the sampled dataset reliably approximate the overall routing behavior across the full dataset for a given task.
This method significantly reduces computational overhead while still capturing representative routing information.
Using the collected routing statistics, \expertune~formulates the expert placement problem as an Integer Linear Programming (ILP) optimization. The ILP optimization integrates the routing data into its constraints and objectives, enabling simultaneous optimization of load balance and communication costs across GPUs.
We use ILP as it offers optimal solutions under the given constraints, ensuring the best expert placement strategies. 
The \expertune~framework operates in two key stages, each modeled as an ILP: (1) Load-Balanced Expert Clustering where we group experts based on routing dependencies while maintaining balanced loads, and (2) Cluster-to-GPU Assignment where these are mapped clusters to GPUs to minimize communication overhead while preserving load balance.
The ILP is formulated using the following:

\begin{itemize}[leftmargin=*]
    \vspace{-0.5em}
    \item \textbf{Modeling Per-GPU Token Processing Load:}
    Using token routing statistics, we quantify the token processing load for each GPU by assigning a weight to each expert proportional to its token demands. This weight reflects the computational workload associated with processing tokens routed to the expert. To ensure balanced token processing, we introduce constraints in the ILP that evenly distribute these loads across GPUs within their capacity limits.
    \vspace{-0.5em}
    \item \textbf{Modeling Inter-GPU Communication Cost:}
    Inter-GPU communication costs are modeled based on token routing frequency between expert pairs. If two interacting experts are placed on separate GPUs, the communication cost is proportional to the volume of tokens transferred between them. This model drives the optimization process, penalizing placements that increase inter-GPU data exchanges and encouraging configurations that minimize cross-GPU communication.
    \vspace{-0.5em}
    \item \textbf{Optimizing expert placement:}
    The expert placement is formulated as two ILP stages where each aims to: (a) balance token processing loads across GPUs and (b) minimize the maximum inter-GPU communication cost. Each ILP incorporates constraints for GPU memory and processing capacity, as well as routing dependencies captured from the token statistics. By solving the optimization problems, we identify an expert placement strategy that maximizes GPU utilization, minimizes idle time, and reduces interconnect communication overhead across all MoE layers.
    \vspace{-0.5em}
\end{itemize}

Through this formulation, \expertune~delivers an efficient and scalable solution for expert placement. By addressing token processing imbalance and inter-GPU communication, it enhances overall system performance, reducing latency and improving throughput in large-scale MoE deployments.

\subsection{\expertune~Overview}~\label{subsec:designoverview}
\vspace{-3ex}

The framework for expert placement in MoE models consists of three stages: Token Routing Profiling, ILP Optimization, and Custom Expert Parallelism Initialization. 
These stages are systematically integrated to optimize expert placement, minimizing token processing imbalance and inter-GPU communication costs. 
The overall process is summarized in Figure ~\ref{fig:methodology-overview}.
In (1) Token Routing Profiling, we analyze token routing patterns by profiling a sampled subset of the dataset, leveraging their consistency across batches to estimate routing dependencies and expert loads. 
Next, in the (2) ILP Optimization stage, we solve the placement problem in two steps: first, we cluster experts to balance token loads across GPUs; then, we determine the optimal mapping of these clusters to GPUs to minimize communication latency induced by token dispatching.
Finally, in the (3) Custom Expert Parallelism Initialization stage, the optimized expert placement is applied to the MoE model, replacing default configurations to improve inference performance by balancing GPU utilization and minimizing communication latency.

\begin{figure}
    \centering
    \includesvg[width=1.0\linewidth]{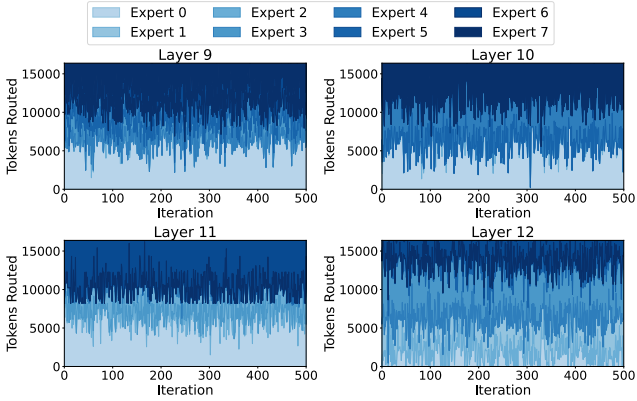}
    \vspace{-1em}
    \caption{Token routing statistics to different experts during the inference of Mixtral-8$\times$7B on the WikiText dataset. The results show that token routing statistics remain consistent across different batches within the same task.}
    \label{fig:routing-invariance}
    \vspace{-2ex}
\end{figure}

\niparagraph{Token Routing Profile}
In this stage, we execute inference over a sampled subset of the task dataset to gather token routing statistics. For each token, we track its routing path across layers in the MoE model. The statistics capture how tokens are routed from one expert to another between neighboring layers.
To better understand the nature of these routing patterns, we analyze their consistency over time. Figure~\ref{fig:routing-invariance} indicates a high degree of invariance in token routing, as illustrated in the accompanying plot. 
This invariance suggests that token routing decisions remain stable across iterations, suggesting that we can use a small subset of the task dataset to optimize the expert placement instead of the entire dataset.
The gathered routing statistics and their consistent patterns will be utilized in the next stage to formulate the expert placement optimization problem, aiming to minimize communication overhead and balance computational loads effectively.

\niparagraph{Leveraging ILP optimization for Expert Placement}
The expert placement in MoE models involves balancing token processing loads and minimizing inter-GPU communication overhead, both of which are highly constrained and interdependent. 
ILP is well-suited because it enables precise modeling of these constraints while optimizing for multiple objectives simultaneously. 
Unlike heuristic methods such as graph partitioning, ILP guarantees globally optimal solutions under the given constraints, ensuring the best expert placement strategies. 
Furthermore, ILP allows for the integration of routing statistics, GPU capacity limitations, and communication costs into a unified optimization problem, making it an ideal choice for solving this complex placement problem.

For expert placement, \expertune~uses the token routing history to construct a routing history table. 
This table captures the number of tokens routed between each pair of neighboring layers in the MoE model. 
The ILP model is formulated with the objective of minimizing load imbalance and inter-GPU communication costs. 
The ILP solver takes routing data, interconnect bandwidth, and resource constraints as inputs and outputs the optimal placement of experts across GPUs. 
After solving the ILP, the optimal expert-to-GPU mapping is saved into a PyTorch tensor file for future use during custom expert parallelism initialization.

\niparagraph{Custom Expert Parallelism}
Using the optimized expert placement derived from the ILP solvers, we initialize the MoE model with the ILP-optimized custom expert parallelism strategy.
The expert-to-GPU mapping file is first loaded from local storage. 
For each layer of the model, the corresponding expert-to-GPU assignments are extracted from the mapping. 
These assignments are then applied to the model to replace the default placement with the optimized configuration.
By ensuring that experts are assigned to GPUs based on the solver’s results, this initialization minimizes communication latency and balances the workload across GPUs.

\subsection{ILP Formulation}~\label{sec:ilp}
%
%
%
The \expertune~optimization comprises two ILPs: ILP 1 for clustering experts to balance token processing loads across GPUs, and ILP 2 for assigning clusters to GPUs while minimizing inter-GPU token routing costs.
%
%
By solving the ILPs, we obtain an expert placement that optimizes GPU and interconnect utilization, enhancing the efficiency of MoE processing by reducing processing load imbalance and communication latency.

\subsubsection{ILP 1: Clustering Experts within Layers}

\niparagraph{Inputs.}
The input to the ILP formulation includes the token routing statistics, which provide information on how many tokens are routed between experts within each MoE layer. 
\begin{itemize}[leftmargin=*]
    \vspace{-0.5em}
    \item \( P_{e,l} \): The number of tokens routed to expert \( e \) in layer \( l \). It indicates the workload associated with each expert and helps calculate the token processing load for each expert cluster during ILP 1.
    \vspace{-0.5em}
    \item \(E\): Number of experts per layer.
    \vspace{-0.5em}
    \item \(L\): Total number of MoE layers.
    \vspace{-0.5em}
    \item \(G\): Total number of GPUs.
\end{itemize}
\niparagraph{Variables.}
The variables of this ILP are:
\vspace{-0.5em}
\begin{align*}
    x_{c,e,l} \in \{0, 1\} \quad \quad \quad &\text{for} \quad c \in \{0, \dots, G-1\}, \\
    &\text{for} \quad e \in \{0, \dots, E-1\}, \\
    &\text{for} \quad \hspace{0.12em} l \in \{0, \dots, L-1\}
\end{align*}
\begin{itemize}
\vspace{-1.5em}
\item \( x_{c,e,l} \): Binary decision variable indicating whether expert \( e \) is assigned to cluster \( c \) in layer \( l \).
\end{itemize}

\niparagraph{Objective Function.}
In the first ILP, the objective is to cluster experts within each layer that balance the token processing load across all clusters, where each cluster will be mapped to a GPU using ILP 2. 
The goal is to distribute the token processing workload as evenly as possible across expert clusters in each layer. 
This is accomplished by minimizing the absolute deviation between the load of each cluster and the average load per layer, ensuring efficient resource utilization. The objective function \(O_1\)can be described as:
\vspace{-0.5ex}
\begin{equation}
    O_1 = \sum_{c=0}^{G-1} \sum_{l=0}^{L-1} \left| T_{c,l} - \bar{T}_l \right|
\end{equation}
\vspace{-0.5ex}
where $T_{c,l}$ is the total token processing load for expert cluster $c$ in layer $l$, and $\bar{T}_l$ is the average token processing load across all clusters in layer $l$.
The token processing load $T_{c,l}$ for each expert cluster in layer $l$ is the sum of token routing statistics for each expert assigned to that cluster:
\vspace{-0.5ex}
\begin{equation}
    T_{c,l} = \sum_{e=0}^{E-1} \sum_{t=0}^{T-1} P_{e,l} \cdot x_{c,e,l}
\end{equation}
\vspace{-0.5ex}
where $P_{e,l}$ is the profiled number of tokens routed to expert $e$ at layer $l$ and $x_{c,e,l}$ is a binary decision variable (1 if expert $e$ is assigned to cluster c , 0 otherwise).
Lastly, the average token processing load across all clusters in a layer is computed as:

\vspace{-0.5ex}
\begin{equation}
    \bar{T}_l = \frac{1}{G} \sum_{e=0}^{E-1} P_{e,l}
\end{equation}
\vspace{-0.5ex}

This objective minimizes the deviation of the token processing load across expert clusters for each layer. 
By minimizing this deviation, we ensure that no cluster is overloaded while others are underutilized.

\niparagraph{Solving the ILP.}
The ILP is executed for every possible expert cluster and for every MoE layer of the model.
The optimization goal is to minimize the deviation in token processing load across expert clusters. The ILP formulation is:
\begin{align}
\text{min} \quad & O_1 \\
\text{s.t.} \quad & O_1 = \sum_{c=0}^{G-1} \sum_{l=0}^{L-1} \left| T_{c,l} - \bar{T}_l \right| \\
\quad & T_{c,l} = \sum_{e=0}^{E-1} P_{e,l} \cdot x_{c,e,l} && (\forall c, l) \\
& \sum_{e=0}^{E-1} x_{c,e,l} \geq 1 && (\forall c, l)
\end{align}

\niparagraph{Constraints.}
The ILP constraints in Equation 7 that at least one expert is assigned to each cluster \( c \) in each layer \( l \), preventing null cluster assignments and ensuring that every layer has sufficient resources.

\subsubsection{ILP 2: Cluster Placement on GPUs}

\niparagraph{Inputs.}
The input to the ILP formulation includes the precomputed communication cost between clusters, which provides information on how many tokens are routed between experts within neighboring MoE layers.
\begin{itemize}[leftmargin=*]
    \vspace{-0.5em}
    \item  \( x_{c,e,l} \) : The binary decision variable indicating whether expert  \( e \)  is assigned to cluster  \( c \)  in layer  \( l \) . This value is determined in ILP 1 and is used to compute the communication cost between clusters for ILP 2.
    \vspace{-0.5em}
    \item \( C_{c_1,c_2,l} \): Number of tokens routed between cluster \( c_1 \) in layer \( l \) and cluster \( c_2 \) in layer \( l+1 \). This represents the number of tokens routed clusters of neighboring layers and is used for balancing the inter-GPU communication load.
    \vspace{-0.5em}
    \item \( R_{e_1, e_2, l} \) : The number of tokens routed between experts \( e_1 \)  and \( e_2 \) in layer \( l \). This value is used to precompute the communication cost between clusters.
    \vspace{-0.5em}
    \item \( B_{g_1,g_2} \): The available bandwidth between GPUs \( g_1 \) and \( g_2 \). This parameter is used to model the bandwidth-aware communication cost by normalizing \( C \) with the available bandwidth, reflecting the relative cost of inter-GPU communication.
    \vspace{-0.5em}
    \item \(E\): Number of experts per layer.
    \vspace{-0.5em}
    \item \(L\): Total number of MoE layers.
    \vspace{-0.5em}
    \item \(G\): Total number of GPUs.
\end{itemize}

\niparagraph{Variables.}
The variables for ILP 2 are defined as follows:
\vspace{-0.5em}
\begin{align*}
    y_{c,g,l} \in \{0, 1\} \quad \quad \quad &\text{for} \quad c \in \{0, \dots, G-1\}, \\
    &\text{for} \quad g \in \{0, \dots, G-1\}, \\
    &\text{for} \quad \hspace{0.12em} l \in \{0, \dots, L-1\}
\end{align*}
\begin{itemize}
\vspace{-0.5em}
\item \( y_{c,g,l} \in \{0, 1\} \) : Binary decision variable indicating whether cluster \( c \) is assigned to GPU \( g \) in layer \( l \) .
\end{itemize}

\niparagraph{Objective Function.}
We aim to minimize the communication overhead between GPUs when routing tokens between experts. 
This is achieved by strategically placing expert clusters on GPUs to reduce inter-GPU communication, with a particular focus on minimizing the all-to-all tail latency in each layer.
The objective function \( O_2 \) directly targets the tail latency per layer by minimizing the maximum communication cost across all GPU pairs, which impacts the latency of token dispatching. The objective function is given by:
\vspace{-1ex}
\begin{equation}
    \small
    O_2 = \sum_{l=0}^{L-1} \max \left( \sum_{c_1, c_2 = 0}^{G-1} \sum_{g_1, g_2 = 0}^{G-1} \frac{C_{c_1, c_2, l}}{B_{g_1, g_2}} \cdot y_{c_1, g_1, l} \cdot y_{c_2, g_2, l+1} \right)
\end{equation}
where $C_{c_1, c_2, l}$ is the communication cost between expert clusters  $c_1$ and $c_2$ for layer $l$ and $l+1$, and $B_{g_1, g_2}$ is the bandwidth between GPUs $g_1$ and $g_2$.
The term $y_{c_1, g_1}$ is a binary decision variable that indicates whether cluster $c_1$ is assigned to GPU $g_1$.
The communication cost $C_{c_1, c_2, l}$ is calculated using the expert assignments from ILP 1, where the $x_{c,e,l}$ values (the binary decision variables indicating whether expert $e$  is assigned to cluster $c$ in layer $l$) have already been determined. 
Using these values, we calculate the total communication cost between expert clusters $c_1$ and $c_2$ for each layer $l$ by summing over all pairs of experts $e_1$ and $e_2$. Specifically, the total communication cost $C_{c_1, c_2, l}$ is computed as:
\vspace{-1ex}
\begin{equation}
    C_{c_1, c_2, l} = \sum_{e_1=0}^{E-1} \sum_{e_2=0}^{E-1} R_{e_1, e_2, l} \cdot x_{c_1,e_1,l} \cdot x_{c_2,e_2,l}
\end{equation}
where $R_{e_1, e_2, l}$ is the number of tokens routed between these experts.
Note that we pre-compute these communication costs \( C_{c_1, c_2, l} \) after ILP 1, thereby avoid having to compute them repeatedly during the ILP 2 optimization process. 

\niparagraph{Solving the ILP.}
The ILP is solved for every possible cluster-GPU mapping across every MoE layer of the model:
\begin{small}
\begin{align}
    \text{min} \quad \quad \quad & O_2 \\
    \text{s.t.} \quad \quad \quad & \notag \\
    O_2 = &\sum_{l=0}^{L-1} \max \left( \sum_{c_1, c_2 = 0}^{G-1} \sum_{g_1, g_2 = 0}^{G-1} \frac{C_{c_1, c_2, l}}{B_{g_1, g_2}} \cdot y_{c_1, g_1, l} \cdot y_{c_2, g_2, l+1} \right) \\
    C_{c_1, c_2, l} = &\sum_{e_1=0}^{E-1} \sum_{e_2=0}^{E-1} R_{e_1, e_2, l} \cdot x_{c_1,e_1,l} \cdot x_{c_2,e_2,l} \quad (\forall c_1, c_2, l) \\
    \quad \sum_{l=0}^{L-1} \sum_{c=0}^{G-1} \sum_{e=0}^{E-1} &x_{c,e,l} \cdot y_{c,g,l} = \frac{E\cdot L}{G} \quad \hspace{6em} (\forall g) \\
    \sum_{g=0}^{G-1} y_{c,g,l} = &1 \quad \hspace{12em} (\forall c, l) \\
    \sum_{c=0}^{G-1} y_{c,g,l} = &1 \quad \hspace{12em} (\forall g, l)
\end{align}
\end{small}

\niparagraph{Constraints.}
The formulation of ILP 2 includes several constraints to ensure a valid and balanced solution. 
These constraints collectively guarantee the one-on-one assignment of clusters to GPUs while respecting the available GPU capacity.
Equation 13 ensures that the total number of experts assigned to each GPU \( g \) across all layers is equal to \( \frac{E \cdot L}{G} \), ensuring a balanced assignment of experts across GPUs. This guarantees a balanced distribution of experts, ensuring that the memory footprint is evenly distributed across GPUs.
Equation 14 enforces that each cluster \( c \) is assigned to exactly one GPU \( g \) in each MoE layer \( l \).
Equation 15 ensures that each GPU \( g \) is assigned to exactly one cluster \( c \) for each layer \( l \), maintaining a balanced distribution of clusters across the GPUs.

\niparagraph{\expertune~Optimization} \expertune~offers a robust solution for optimizing expert placement in large-scale MoE models.
By leveraging routing dependencies and systematically balancing token processing loads, while minimizing inter-GPU communication costs, \expertune~achieves significant improvements in both GPU utilization and overall system throughput.
Next, we analyze the performance of \expertune~across diverse configurations and datasets, demonstrating its capability to address key challenges in MoE inference effectively.

%% file: body/evaluation.tex
\section{Evaluation}
\subsection{Experimental Setup}~\label{subsec:setup}
\vspace{-2ex}

\niparagraph{MoE Model and Datasets.} We evaluate \expertune~on pre-trained Mixtral8x7B~\cite{mixtral} available on the Huggingface Hub~\cite{huggingface}, benchmarking its performance on a representative selection of language modeling datasets, as shown in Table~\ref{tab:evaluation-datasets}. 
As discussed in Section~\ref{subsec:designoverview}, the routing patterns observed with the subset closely match those of the full datasets, providing reliable insights into \expertune's performance. 

\begin{table}
    \centering
    \caption{Evaluation Datasets}
    \vspace{-0.7em}
    \resizebox{0.8\columnwidth}{!}{
    \begin{tabular}{|l|l|l|}
        \hline
        \textbf{Dataset} & \textbf{Abbreviation} & \textbf{Type} \\
        \hline
        WikiText-103~\cite{wikitext} & wiki & Language Modeling \\
        \hline
        MiniPile~\cite{minipile} & pile & Language Modeling \\
        \hline
        LAMBADA~\cite{lambada} & lamb & Language Modeling \\
        \hline
        enwik8~\cite{enwik8} & enwi & Language Modeling \\
        \hline
    \end{tabular}
    }
    \label{tab:evaluation-datasets}
\end{table}

\niparagraph{Expert and Tensor Parallel Configurations.}
Since Mixtral features eight experts per layer, we limit the size of expert parallelism (EP) to four. 
Our methodology remains broadly applicable as the number of experts is expected to scale further in the future to accommodate greater knowledge capacity. To scale beyond four GPUs, we employ a hybrid parallelism strategy combining tensor parallelism (TP) and expert parallelism.
For single-node (8 GPUs) and multi-node (2 nodes and 16 GPUs total) experiments, we configured parallelism as 4EP-2TP and 4EP-4TP, respectively.

\niparagraph{Software and Libraries and Setup.}
To implement \expertune, we modify the all-to-all communication and expert placement modules in Megatron-LM~\cite{megatron} to allow custom expert mappings across GPUs, supporting variable numbers of experts per layer. Our ILP was optimized using Gurobi~\cite{gurobi} (version 12.0.0). Both ILPs were set to execute until reaching a tolerance of 0.025, meaning the solver iteratively refines the solution by adjusting the values of decision variables to minimize the objective function. The evaluation was conducted with PyTorch 2.5.1~\cite{pytorch}, CUDA Toolkit 12.4~\cite{cuda,nccl}, and RHEL 9 OS~\cite{rhel9}.

\begin{table}
    \centering
    \caption{H100 Server Node Specifications}
    \vspace{-0.7em}
    \resizebox{0.8\columnwidth}{!}{
    \begin{tabular}{|l|l|}
        \hline
        \textbf{Component} & \textbf{Specification} \\
        \hline
        GPU & 8x NVIDIA H100 SXM5 80GB \\
        \hline
        Interconnect & NVLink Gen4 (900GB/s) \\
        \hline
        CPU & Dual Xeon Platinum 8462Y+ \\
        \hline
        System Memory & 2048 GB DDR5 4800 MHz \\
        \hline
        NIC & NVIDIA ConnectX-7 IB (400Gbps) \\
        \hline
    \end{tabular}}
    \label{tab:hardware-specs}
\end{table}

\niparagraph{Server Architecture.}
Experiments were conducted on a high-performance computing node with specifications detailed in Table~\ref{tab:hardware-specs} and Table~\ref{tab:hardware-specs1}. Our evaluations were performed in both single-node (i.e., 1 node with 8 GPUs) and multi-node (i.e., 2 nodes with 8 GPUs each) configurations to assess the impact of hierarchical interconnect topologies, especially in terms of how communication patterns change across nodes. Due to resource availability, single-node experiments were conducted using NVIDIA H100 GPUs~\cite{h100} while multi-node experiments were performed on NVIDIA H200 GPUs~\cite{h200}.

\begin{table}
    \centering
    \caption{H200 Server Node Specifications}
    \vspace{-0.7em}
    \resizebox{0.8\columnwidth}{!}{
    \begin{tabular}{|l|l|}
        \hline
        \textbf{Component} & \textbf{Specification} \\
        \hline
        GPU & 8x NVIDIA H200 SXM5 142GB \\
        \hline
        Interconnect & NVLink Gen4 (900GB/s) \\
        \hline
        CPU & Dual Xeon Platinum 8562Y \\
        \hline
        System Memory & 2048 GB DDR5 5600 MHz \\
        \hline
        NIC & NVIDIA ConnectX-7 IB (800Gbps) \\
        \hline
    \end{tabular}}
    \label{tab:hardware-specs1}
\end{table}
\vspace{-1em}

\subsection{Baseline and Metrics}

\niparagraph{Baseline and Expert Assignment.}  
We use Megatron-LM~\cite{megatron} as the baseline, which employs a naive expert placement strategy where experts are assigned to GPUs in contiguous blocks (for example, with 8 experts and 4 GPUs, experts 0 and 1 are assigned to GPU 0, experts 2 and 3 to GPU 1, and so on). 
This approach ensures an even distribution of experts for memory footprint but does not address load balancing or communication inefficiencies during runtime. 
%

\niparagraph{Evaluation Metrics.} 
We evaluate \expertune and the baseline using the following metrics: \textit{End-to-End Speedup:} Speedup in absolute time to complete one batch of inference. \textit{Token Processing Time:} Tail latency and average time taken for experts to finish processing tokens at each layer. \textit{All-to-All Time:} Tail latency and average time taken to complete all-to-all communication across GPUs.

\niparagraph{End-to-End Speedup Measurement.}
End-to-end speedup is measured by running 100 inference steps and averaging the results across datasets. To ensure stable measurements, 100 warmup steps are performed prior to recording the timing.

\niparagraph{Tail- and Average- Latency Measurement.}
For all latency measurements, we employed PyTorch Profiler~\cite{torchprof}, averaging results over 10 inference steps with an initial 100 warm-up steps on the WikiText-103 dataset.
To measure tail latency, we identify the GPU with the longest execution time at each layer for each iteration, then calculate the average of these maximum values across all iterations. 
For average latency, we compute the mean execution time across all GPUs for each iteration and then average these means across iterations.

\subsection{Results and Insights}

\begin{figure}
    \centering
    \includegraphics[width=1.0\linewidth]{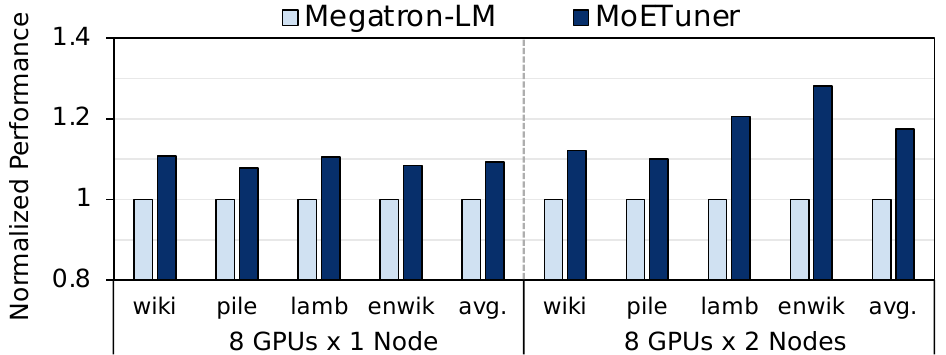}
    \vspace{-1.5em}
    \caption{End-to-end inference performance, normalized to Megatron-LM's expert parallelism approach.}
    \label{fig:speedup}
\end{figure}

\subsubsection{End-to-End Speedup}
\begin{figure}
    \centering
    \includesvg[width=1\linewidth]{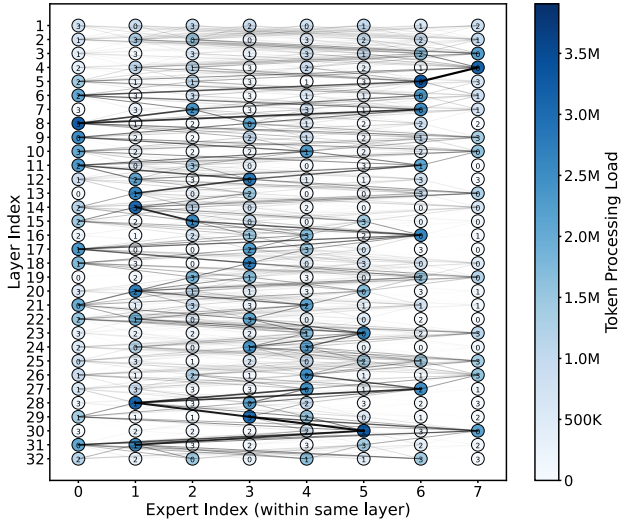}
    \vspace{-1em}
    \caption{Token routing statistics for Mixtral-8x7B, with custom mapping generated by \expertune. The circle represents the index of expert parallel rank assignment.}
    \label{fig:mapping}
\end{figure}

Figure~\ref{fig:speedup} compares the end-to-end speedup of \expertune~against the baseline Megatron-LM in both single-node and multi-node settings.
\expertune~achieves a 9.3\% speedup in the single-node setup and a 17.5\% speedup in the multi-node setup.
These performance improvements are driven by reductions in communication overhead and efficient load balancing of token processing across GPUs.
Figure~\ref{fig:mapping} illustrates the token routing statistics of Mixtral-8x7B on the WikiText-103 dataset, with the mapping generated by \expertune. 
\expertune~demonstrates significant improvements in token load balancing, as evidenced by the darker expert colors indicating high token load on GPUs with only one expert assigned per layer, while lighter expert colors indicate low token load on GPUs with a larger number of experts.
Moreover, \expertune~also effectively manages remote token dispatching, as observed in layers 4 and 5, where experts 7 and 6 are mapped to the same GPU.
In the single-node setup, the speedup can primarily be attributed to better resource utilization, with load balancing minimizing GPU idle times.
Due to the balanced token processing across GPUs, \expertune~enhances GPU memory utilization by 8.7\% in the single-node setup.

\begin{figure}
    \centering
    \begin{subfigure}[b]{\linewidth}
        \centering
        \includesvg[width=1\linewidth]{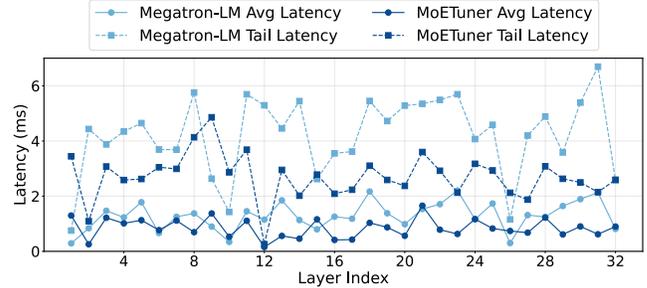}  
        \vspace{-1.8em}
        \caption{Single node: Average and tail latency of token processing time across GPUs in each layer.}
        \vspace{0.5em}
        \label{fig:tail-latency-computation-1node}
    \end{subfigure}
    \begin{subfigure}[b]{\linewidth}
        \centering
        \includesvg[width=1\linewidth]{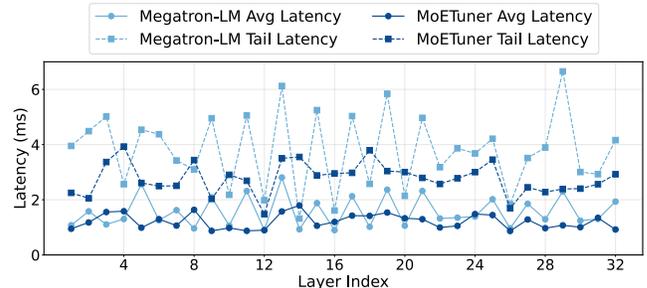}  
        \vspace{-1.8em}
        \caption{Multi node: Average and tail latency of token processing time across GPUs in each layer.}
        \label{fig:tail-latency-computation-2nodes}
    \end{subfigure}
    \vspace{-2em}
    \caption{Comparison of average and tail latency of token processing time across GPUs in each layer for single-node and multi-node setups.}
    \label{fig:compute-tail-latency-comparison}
\end{figure}

In the multi-node setup, reduced inter-node communication overhead leads to a more significant speedup by minimizing communication volume, which is especially costly in multi-node configurations.
This is due to the much lower inter-node network bandwidth, 9$\times$ lower than intra-node NVLink bandwidth.
The lower inter-node bandwidth results in higher all-to-all communication latencies, but \expertune~effectively reduces inter-node communication volume, leading to a more substantial speedup compared to the single-node scenario.
As a result, our dual approach-targeting both computation and communication tail latency-enables consistent performance gains in both single-node and multi-node scenarios.
As such this shows the benefits of \expertune~and its scalability to multi-node setting.
In the next part of this section, we break down the reasons for this performance benefit through token processing and communication time reduction.

\begin{figure}[h!]
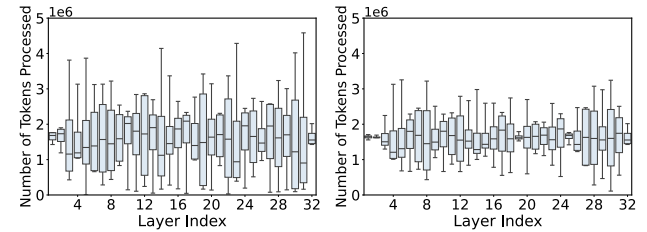

    \centering
    \begin{subfigure}[b]{0.485\linewidth}
        \centering
        \includesvg[width=1\linewidth]{figures/processing-load-vanilla.svg}
        \caption{Token processing load distribution (Megatron-LM).}
        \label{fig:processing-load-baseline}
    \end{subfigure}
    \begin{subfigure}[b]{0.485\linewidth}
        \centering
        \includesvg[width=1\linewidth]{figures/processing-load-custom.svg}
        \caption{Token processing load distribution (\expertune).}
        \label{fig:processing-load-expertune}
    \end{subfigure}
    \caption{Distribution of tokens processed by a single GPU per layer. Each box plot summarizes the variation in token processing load across GPUs for a single layer. \expertune~significantly reduces both the variation and peak load, demonstrating improved load balancing across GPUs.}
    \label{fig:processing-load}
\end{figure}

\subsubsection{Token Processing Time}
As shown in Figure~\ref{fig:compute-tail-latency-comparison}, \expertune~reduces the tail latency in token processing 36\% in the single-node setup and by 27\% in the multi-node setup.
Similarly, the average token processing time is reduced by 34.8\% in the single-node configuration and 22.5\% in the multi-node configuration.
These improvements are achieved through improved token dispatching and load balancing across GPUs, which minimizes stragglers during token processing.
The benefits are particularly pronounced in layers with high computational loads, such as layer 31, where \expertune~significantly reduces both average and tail processing times. 
This is likely due to its capacity to mitigate imbalances in token dispatching, as evidenced by the token routing distributions in Figure~\ref{fig:processing-load}. 
By preventing GPUs from being overloaded or underutilized, \expertune~ensures more consistent and efficient computation, even in the most demanding layers.

\begin{figure}
    \centering
    \begin{subfigure}[b]{\linewidth}
        \centering
        \includesvg[width=1\linewidth]{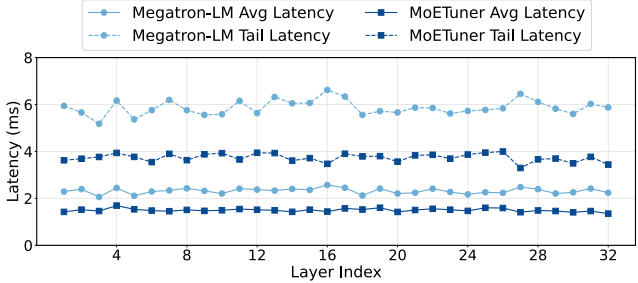}  
        \vspace{-1.8em}
        \caption{Single node: Average and tail latency of all-to-all time across GPUs in each neighboring layers.}
        \vspace{0.5em}
        \label{fig:tail-latency-communication-1node}
    \end{subfigure}
    \begin{subfigure}[b]{\linewidth}
        \centering
        \includesvg[width=1\linewidth]{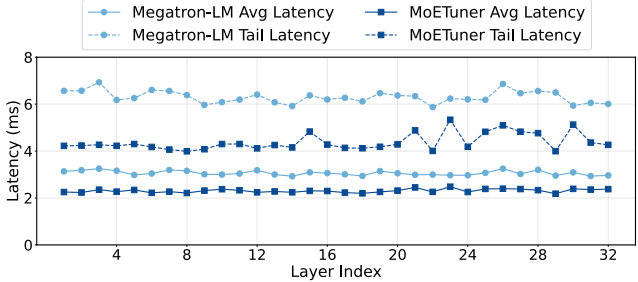}  
        \vspace{-1.8em}
        \caption{Multi node: Average and tail latency of token processing time across GPUs in each neighboring layers.}
        \label{fig:tail-latency-communication-2nodes}
    \end{subfigure}
    \vspace{-2em}
    \caption{Comparison of average and tail latency of all-to-all communication in each layer for single-node and multi-node setups. \expertune~provides substantial reduction in tail latency and average latency.}
    \label{fig:communication-tail-latency-comparison}
\end{figure}

In the multi-node setup, while token processing time is reduced, the improvements are less pronounced compared to the single-node configuration. 
This is primarily due to the additional inter-node communication overhead during token processing, specifically the costs of all-gather operations introduced by tensor parallelism, as demonstrated in Figure~\ref{fig:time-distribution}.
As discussed in Section~\ref{subsec:setup}, our experimental design keeps the expert parallel size fixed while scaling the tensor parallel size, which increases the all-gather communication volume required to synchronize intermediate activations.
Although some layers, such as those in the 12–22 range, exhibit fluctuations in latency, this variability is likely due to the short profiling iterations.
Despite these anomalies, \expertune~demonstrates consistent benefits across setups, effectively addressing inefficiencies in token processing to deliver predictable and efficient computation.

\begin{figure}
    \centering
    \begin{subfigure}[b]{0.485\linewidth}
        \centering
        \includesvg[width=1\linewidth]{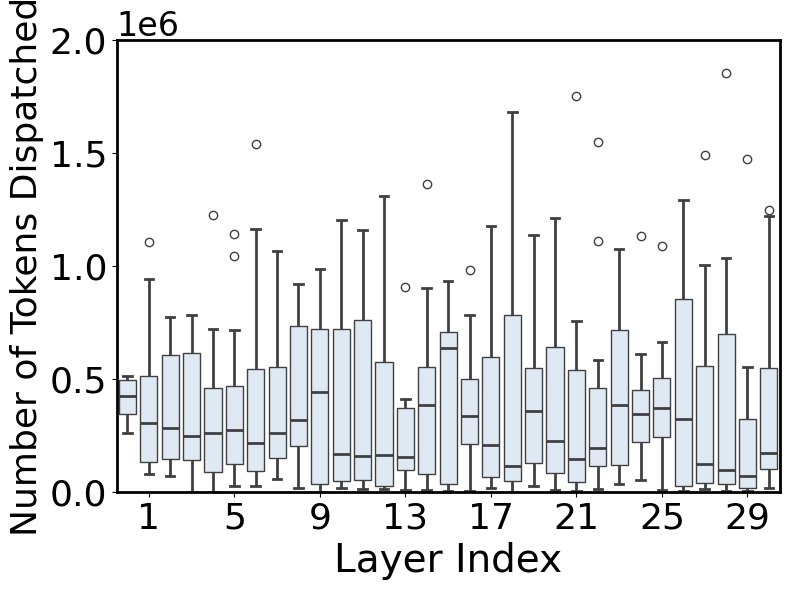}
        \caption{Token dispatching distribution (Megatron-LM).}
        \label{fig:comm-vanilla}
    \end{subfigure}
    \begin{subfigure}[b]{0.485\linewidth}
        \centering
        \includesvg[width=1\linewidth]{figures/comm-volume-custom.svg}
        \caption{Token dispatching distribution (\expertune).}
        \label{fig:comm-custom}
    \end{subfigure}
    \caption{Distribution of total token dispatching between individual GPU pairs across neighboring layers, measured for a single-node configuration. Each data point in a box plot represents the total number of tokens dispatched between a specific GPU pair (e.g., GPU0-GPU1) across all iterations.
    }
    \vspace{-1em}
    \label{fig:comm-volume}
\end{figure}

\subsubsection{All-to-All Time}
As illustrated in Figure~\ref{fig:communication-tail-latency-comparison}, our proposed optimization strategy significantly reduces the tail latency of all-to-all communication, which is critical for improving end-to-end inference time.
In the single-node setup, \expertune~reduces the tail latency by 36.3\% and the average latency by 35.4\% compared to the baseline.
This reduction is consistent across most layers, with notable improvements in layers 13-17, where both average and tail latencies are particularly low.
In the multi-node setup, \expertune~achieves a 30.50\% reduction in tail latency and a 24.7\% reduction in average latency.
The reduced impact in this setup is primarily due to the higher inter-node communication overhead, which dominates the latency in multi-node environments.
For example, in certain layers (e.g., layers 15 and 30), the improvement in tail latency is less pronounced.
This is attributed to high inter-node token dispatching in these layers, which introduces additional communication overhead.
The tail latency spikes observed in such layers are consistent with the token dispatching distributions shown in Figure~\ref{fig:comm-volume}, where layers 15 and 30 exhibit relatively higher maximum communication volumes compared to neighboring layers.
Despite these spikes, \expertune~still provides measurable benefits over the baseline, maintaining lower latency overall.
By reducing the variation in token dispatching and avoiding severe imbalances, our approach ensures more efficient and predictable all-to-all communication.

%% file: body/conclusion.tex
\section{Conclusion}
We present \expertune, a method that optimizes MoE models by enhancing token dispatching and load balancing across GPUs.
Our experiments on Mixtral-8x7B demonstrated significant reductions in both tail latency and average token processing time, particularly in layers with high token routing skew. 
Furthermore, \expertune~mitigates the impact of inter-GPU communication overhead by balancing remote token dispatching and ensuring efficient all-to-all communication.

%% file: main.bbl
\begin{thebibliography}{10}

\bibitem{gpt3}
Tom~B. Brown, Benjamin Mann, Nick Ryder, Melanie Subbiah, Jared Kaplan, Prafulla Dhariwal, Arvind Neelakantan, Pranav Shyam, Girish Sastry, Amanda Askell, Sandhini Agarwal, Ariel Herbert-Voss, Gretchen Krueger, Tom Henighan, Rewon Child, Aditya Ramesh, Daniel~M. Ziegler, Jeffrey Wu, Clemens Winter, Christopher Hesse, Mark Chen, Eric Sigler, Mateusz Litwin, Scott Gray, Benjamin Chess, Jack Clark, Christopher Berner, Sam McCandlish, Alec Radford, Ilya Sutskever, and Dario Amodei.
\newblock Language models are few-shot learners, 2020.

\bibitem{cai2024shortcut}
Weilin Cai, Juyong Jiang, Le~Qin, Junwei Cui, Sunghun Kim, and Jiayi Huang.
\newblock Shortcut-connected expert parallelism for accelerating mixture-of-experts.
\newblock {\em arXiv preprint arXiv:2404.05019}, 2024.

\bibitem{cai2024survey}
Weilin Cai, Juyong Jiang, Fan Wang, Jing Tang, Sunghun Kim, and Jiayi Huang.
\newblock A survey on mixture of experts.
\newblock {\em arXiv preprint arXiv:2407.06204}, 2024.

\bibitem{chen2022ta}
Chang Chen, Min Li, Zhihua Wu, Dianhai Yu, and Chao Yang.
\newblock Ta-moe: Topology-aware large scale mixture-of-expert training.
\newblock {\em Advances in Neural Information Processing Systems}, 35:22173--22186, 2022.

\bibitem{switchtransformer}
William Fedus, Barret Zoph, and Noam Shazeer.
\newblock Switch transformers: scaling to trillion parameter models with simple and efficient sparsity.
\newblock {\em J. Mach. Learn. Res.}, 23(1), January 2022.

\bibitem{gurobi}
{Gurobi Optimization, LLC}.
\newblock {Gurobi Optimizer Reference Manual}, 2024.
\newblock \url{https://www.gurobi.com}.

\bibitem{fastmoe}
Jiaao He, Jiezhong Qiu, Aohan Zeng, Zhilin Yang, Jidong Zhai, and Jie Tang.
\newblock Fastmoe: A fast mixture-of-expert training system.
\newblock {\em arXiv preprint arXiv:2103.13262}, 2021.

\bibitem{fastermoe}
Jiaao He, Jidong Zhai, Tiago Antunes, Haojie Wang, Fuwen Luo, Shangfeng Shi, and Qin Li.
\newblock Fastermoe: modeling and optimizing training of large-scale dynamic pre-trained models.
\newblock In {\em Proceedings of the 27th ACM SIGPLAN Symposium on Principles and Practice of Parallel Programming}, pages 120--134, 2022.

\bibitem{huang2023towards}
Haiyang Huang, Newsha Ardalani, Anna Sun, Liu Ke, Hsien-Hsin~S Lee, Anjali Sridhar, Shruti Bhosale, Carole-Jean Wu, and Benjamin Lee.
\newblock Towards moe deployment: Mitigating inefficiencies in mixture-of-expert (moe) inference.
\newblock {\em arXiv preprint arXiv:2303.06182}, 2023.

\bibitem{huang2019gpipe}
Yanping Huang, Youlong Cheng, Ankur Bapna, Orhan Firat, Dehao Chen, Mia Chen, HyoukJoong Lee, Jiquan Ngiam, Quoc~V Le, Yonghui Wu, et~al.
\newblock Gpipe: Efficient training of giant neural networks using pipeline parallelism.
\newblock {\em Advances in neural information processing systems}, 32, 2019.

\bibitem{huggingface}
Huggingface.
\newblock {Huggingface Hub documentation}, 2024.
\newblock \url{https://huggingface.co/docs/hub/en/index}.

\bibitem{enwik8}
Marcus Hutter.
\newblock The human knowledge compression contest, 2006.
\newblock \url{http://prize.hutter1.net}.

\bibitem{hwang2023tutel}
Changho Hwang, Wei Cui, Yifan Xiong, Ziyue Yang, Ze~Liu, Han Hu, Zilong Wang, Rafael Salas, Jithin Jose, Prabhat Ram, et~al.
\newblock Tutel: Adaptive mixture-of-experts at scale.
\newblock {\em Proceedings of Machine Learning and Systems}, 5:269--287, 2023.

\bibitem{hwang2024pre}
Ranggi Hwang, Jianyu Wei, Shijie Cao, Changho Hwang, Xiaohu Tang, Ting Cao, and Mao Yang.
\newblock Pre-gated moe: An algorithm-system co-design for fast and scalable mixture-of-expert inference.
\newblock In {\em 2024 ACM/IEEE 51st Annual International Symposium on Computer Architecture (ISCA)}, pages 1018--1031. IEEE, 2024.

\bibitem{mixtral}
Albert~Q. Jiang, Alexandre Sablayrolles, Antoine Roux, Arthur Mensch, Blanche Savary, Chris Bamford, Devendra~Singh Chaplot, Diego de~las Casas, Emma~Bou Hanna, Florian Bressand, Gianna Lengyel, Guillaume Bour, Guillaume Lample, Lélio~Renard Lavaud, Lucile Saulnier, Marie-Anne Lachaux, Pierre Stock, Sandeep Subramanian, Sophia Yang, Szymon Antoniak, Teven~Le Scao, Théophile Gervet, Thibaut Lavril, Thomas Wang, Timothée Lacroix, and William~El Sayed.
\newblock Mixtral of experts, 2024.
\newblock \url{https://arxiv.org/abs/2401.04088}.

\bibitem{jiang2024lancet}
Chenyu Jiang, Ye~Tian, Zhen Jia, Shuai Zheng, Chuan Wu, and Yida Wang.
\newblock Lancet: Accelerating mixture-of-experts training via whole graph computation-communication overlapping.
\newblock {\em arXiv preprint arXiv:2404.19429}, 2024.

\bibitem{minipile}
Jean Kaddour.
\newblock The minipile challenge for data-efficient language models.
\newblock {\em arXiv preprint arXiv:2304.08442}, 2023.

\bibitem{kim2021scalable}
Young~Jin Kim, Ammar~Ahmad Awan, Alexandre Muzio, Andres Felipe~Cruz Salinas, Liyang Lu, Amr Hendy, Samyam Rajbhandari, Yuxiong He, and Hany~Hassan Awadalla.
\newblock Scalable and efficient moe training for multitask multilingual models.
\newblock {\em arXiv preprint arXiv:2109.10465}, 2021.

\bibitem{vllm}
Woosuk Kwon, Zhuohan Li, Siyuan Zhuang, Ying Sheng, Lianmin Zheng, Cody~Hao Yu, Joseph~E. Gonzalez, Hao Zhang, and Ion Stoica.
\newblock Efficient memory management for large language model serving with pagedattention.
\newblock In {\em Proceedings of the ACM SIGOPS 29th Symposium on Operating Systems Principles}, 2023.

\bibitem{gshard}
Dmitry Lepikhin, HyoukJoong Lee, Yuanzhong Xu, Dehao Chen, Orhan Firat, Yanping Huang, Maxim Krikun, Noam Shazeer, and Zhifeng Chen.
\newblock Gshard: Scaling giant models with conditional computation and automatic sharding.
\newblock {\em arXiv preprint arXiv:2006.16668}, 2020.

\bibitem{li2023accelerating}
Jiamin Li, Yimin Jiang, Yibo Zhu, Cong Wang, and Hong Xu.
\newblock Accelerating distributed $\{$MoE$\}$ training and inference with lina.
\newblock In {\em 2023 USENIX Annual Technical Conference (USENIX ATC 23)}, pages 945--959, 2023.

\bibitem{mcsmoe}
Pingzhi Li, Zhenyu Zhang, Prateek Yadav, Yi-Lin Sung, Yu~Cheng, Mohit Bansal, and Tianlong Chen.
\newblock Merge, then compress: Demystify efficient smoe with hints from its routing policy.
\newblock {\em arXiv preprint arXiv:2310.01334}, 2023.

\bibitem{li2020pytorch}
Shen Li, Yanli Zhao, Rohan Varma, Omkar Salpekar, Pieter Noordhuis, Teng Li, Adam Paszke, Jeff Smith, Brian Vaughan, Pritam Damania, et~al.
\newblock Pytorch distributed: Experiences on accelerating data parallel training.
\newblock {\em arXiv preprint arXiv:2006.15704}, 2020.

\bibitem{wikitext}
Stephen Merity, Caiming Xiong, James Bradbury, and Richard Socher.
\newblock Pointer sentinel mixture models, 2016.

\bibitem{narayanan2019pipedream}
Deepak Narayanan, Aaron Harlap, Amar Phanishayee, Vivek Seshadri, Nikhil~R Devanur, Gregory~R Ganger, Phillip~B Gibbons, and Matei Zaharia.
\newblock Pipedream: Generalized pipeline parallelism for dnn training.
\newblock In {\em Proceedings of the 27th ACM symposium on operating systems principles}, pages 1--15, 2019.

\bibitem{narayanan2021memory}
Deepak Narayanan, Amar Phanishayee, Kaiyu Shi, Xie Chen, and Matei Zaharia.
\newblock Memory-efficient pipeline-parallel dnn training.
\newblock In {\em International Conference on Machine Learning}, pages 7937--7947. PMLR, 2021.

\bibitem{h100}
NVIDIA.
\newblock {NVIDIA H100 Tensor Core GPU}, 2023.
\newblock \url{https://resources.nvidia.com/en-us-tensor-core/nvidia-tensor-core-gpu-datasheet}.

\bibitem{cuda}
NVIDIA.
\newblock {CUDA Toolkit}, 2024.
\newblock \url{https://developer.nvidia.com/cuda-toolkit}.

\bibitem{nccl}
NVIDIA.
\newblock {NVIDIA Collective Communications Library (NCCL)}, 2024.
\newblock \url{https://developer.nvidia.com/nccl}.

\bibitem{h200}
NVIDIA.
\newblock {NVIDIA H200 Tensor Core GPU}, 2024.
\newblock \url{https://resources.nvidia.com/en-us-data-center-overview-mc/en-us-data-center-overview/hpc-datasheet-sc23-h200}.

\bibitem{lambada}
Denis Paperno, Germ{\'a}n Kruszewski, Angeliki Lazaridou, Quan~Ngoc Pham, Raffaella Bernardi, Sandro Pezzelle, Marco Baroni, Gemma Boleda, and Raquel Fern{\'a}ndez.
\newblock The lambada dataset: Word prediction requiring a broad discourse context.
\newblock {\em arXiv preprint arXiv:1606.06031}, 2016.

\bibitem{pytorch}
Adam Paszke, Sam Gross, Francisco Massa, Adam Lerer, James Bradbury, Gregory Chanan, Trevor Killeen, Zeming Lin, Natalia Gimelshein, Luca Antiga, et~al.
\newblock Pytorch: An imperative style, high-performance deep learning library.
\newblock {\em Advances in neural information processing systems}, 32, 2019.

\bibitem{torchprof}
PyTorch.
\newblock {PyTorch Profiler}, February 2023.
\newblock \url{https://pytorch.org/tutorials/recipes/recipes/profiler_recipe.html}.

\bibitem{gpt2}
Alec Radford, Jeffrey Wu, Rewon Child, David Luan, Dario Amodei, Ilya Sutskever, et~al.
\newblock Language models are unsupervised multitask learners.
\newblock {\em OpenAI blog}, 1(8):9, 2019.

\bibitem{t5}
Colin Raffel, Noam Shazeer, Adam Roberts, Katherine Lee, Sharan Narang, Michael Matena, Yanqi Zhou, Wei Li, and Peter~J Liu.
\newblock Exploring the limits of transfer learning with a unified text-to-text transformer.
\newblock {\em Journal of machine learning research}, 21(140):1--67, 2020.

\bibitem{deepspeed-moe}
Samyam Rajbhandari, Conglong Li, Zhewei Yao, Minjia Zhang, Reza~Yazdani Aminabadi, Ammar~Ahmad Awan, Jeff Rasley, and Yuxiong He.
\newblock Deepspeed-moe: Advancing mixture-of-experts inference and training to power next-generation ai scale.
\newblock In {\em International conference on machine learning}, pages 18332--18346. PMLR, 2022.

\bibitem{deepspeed}
Jeff Rasley, Samyam Rajbhandari, Olatunji Ruwase, and Yuxiong He.
\newblock Deepspeed: System optimizations enable training deep learning models with over 100 billion parameters.
\newblock In {\em Proceedings of the 26th ACM SIGKDD International Conference on Knowledge Discovery \& Data Mining}, pages 3505--3506, 2020.

\bibitem{rhel9}
{Red Hat, Inc.}
\newblock {Red Hat Enterprise Linux 9}, 2024.
\newblock \url{https://docs.redhat.com/en/documentation/red_hat_enterprise_linux/9}.

\bibitem{shazeer2017outrageously}
Noam Shazeer, Azalia Mirhoseini, Krzysztof Maziarz, Andy Davis, Quoc Le, Geoffrey Hinton, and Jeff Dean.
\newblock Outrageously large neural networks: The sparsely-gated mixture-of-experts layer.
\newblock {\em arXiv preprint arXiv:1701.06538}, 2017.

\bibitem{megatron}
Mohammad Shoeybi, Mostofa Patwary, Raul Puri, Patrick LeGresley, Jared Casper, and Bryan Catanzaro.
\newblock Megatron-lm: Training multi-billion parameter language models using model parallelism.
\newblock {\em arXiv preprint arXiv:1909.08053}, 2019.

\bibitem{singh2023hybrid}
Siddharth Singh, Olatunji Ruwase, Ammar~Ahmad Awan, Samyam Rajbhandari, Yuxiong He, and Abhinav Bhatele.
\newblock A hybrid tensor-expert-data parallelism approach to optimize mixture-of-experts training.
\newblock In {\em Proceedings of the 37th International Conference on Supercomputing}, pages 203--214, 2023.

\bibitem{gemma}
Gemma Team, Thomas Mesnard, Cassidy Hardin, Robert Dadashi, Surya Bhupatiraju, Shreya Pathak, Laurent Sifre, Morgane Rivière, Mihir~Sanjay Kale, Juliette Love, Pouya Tafti, Léonard Hussenot, Pier~Giuseppe Sessa, Aakanksha Chowdhery, Adam Roberts, Aditya Barua, Alex Botev, Alex Castro-Ros, Ambrose Slone, Amélie Héliou, Andrea Tacchetti, Anna Bulanova, Antonia Paterson, Beth Tsai, Bobak Shahriari, Charline~Le Lan, Christopher~A. Choquette-Choo, Clément Crepy, Daniel Cer, Daphne Ippolito, David Reid, Elena Buchatskaya, Eric Ni, Eric Noland, Geng Yan, George Tucker, George-Christian Muraru, Grigory Rozhdestvenskiy, Henryk Michalewski, Ian Tenney, Ivan Grishchenko, Jacob Austin, James Keeling, Jane Labanowski, Jean-Baptiste Lespiau, Jeff Stanway, Jenny Brennan, Jeremy Chen, Johan Ferret, Justin Chiu, Justin Mao-Jones, Katherine Lee, Kathy Yu, Katie Millican, Lars~Lowe Sjoesund, Lisa Lee, Lucas Dixon, Machel Reid, Maciej Mikuła, Mateo Wirth, Michael Sharman, Nikolai Chinaev, Nithum Thain, Olivier Bachem,
  Oscar Chang, Oscar Wahltinez, Paige Bailey, Paul Michel, Petko Yotov, Rahma Chaabouni, Ramona Comanescu, Reena Jana, Rohan Anil, Ross McIlroy, Ruibo Liu, Ryan Mullins, Samuel~L Smith, Sebastian Borgeaud, Sertan Girgin, Sholto Douglas, Shree Pandya, Siamak Shakeri, Soham De, Ted Klimenko, Tom Hennigan, Vlad Feinberg, Wojciech Stokowiec, Yu~hui Chen, Zafarali Ahmed, Zhitao Gong, Tris Warkentin, Ludovic Peran, Minh Giang, Clément Farabet, Oriol Vinyals, Jeff Dean, Koray Kavukcuoglu, Demis Hassabis, Zoubin Ghahramani, Douglas Eck, Joelle Barral, Fernando Pereira, Eli Collins, Armand Joulin, Noah Fiedel, Evan Senter, Alek Andreev, and Kathleen Kenealy.
\newblock Gemma: Open models based on gemini research and technology, 2024.
\newblock \url{https://arxiv.org/abs/2403.08295}.

\bibitem{llama}
Hugo Touvron, Thibaut Lavril, Gautier Izacard, Xavier Martinet, Marie-Anne Lachaux, Timothée Lacroix, Baptiste Rozière, Naman Goyal, Eric Hambro, Faisal Azhar, Aurelien Rodriguez, Armand Joulin, Edouard Grave, and Guillaume Lample.
\newblock Llama: Open and efficient foundation language models, 2023.
\newblock \url{https://arxiv.org/abs/2302.13971}.

\bibitem{bert}
A~Vaswani.
\newblock Attention is all you need.
\newblock {\em Advances in Neural Information Processing Systems}, 2017.

\bibitem{exflow}
Jinghan Yao, Quentin Anthony, Aamir Shafi, Hari Subramoni, and Dhabaleswar K~DK Panda.
\newblock Exploiting inter-layer expert affinity for accelerating mixture-of-experts model inference.
\newblock In {\em 2024 IEEE International Parallel and Distributed Processing Symposium (IPDPS)}, pages 915--925. IEEE, 2024.

\bibitem{zhao2023pytorch}
Yanli Zhao, Andrew Gu, Rohan Varma, Liang Luo, Chien-Chin Huang, Min Xu, Less Wright, Hamid Shojanazeri, Myle Ott, Sam Shleifer, et~al.
\newblock Pytorch fsdp: experiences on scaling fully sharded data parallel.
\newblock {\em arXiv preprint arXiv:2304.11277}, 2023.

\end{thebibliography}
